\documentclass[letterpaper, 10 pt, conference]{ieeeconf}  

\IEEEoverridecommandlockouts                              

\usepackage{graphics} 
\usepackage{epsfig} 
\usepackage{mathptmx} 
\usepackage{times} 
\usepackage{amsmath} 
\usepackage{amssymb}  
\usepackage{cite}
\usepackage{xcolor}
\usepackage{subfigure}
\RequirePackage{multicol}
\RequirePackage{multirow}
\usepackage{rotating}
\usepackage{makecell}
\usepackage{hyperref}
\usepackage{booktabs}
\usepackage{graphicx}
\RequirePackage{epstopdf}
\usepackage{xcolor,colortbl}

\title{\LARGE \bf
EnvoDat: A Large-Scale Multisensory Dataset for Robotic Spatial Awareness and Semantic Reasoning in Heterogeneous Environments
}

\author{
Linus Nwankwo$^{1*}$, Björn Ellensohn$^{1}$, Vedant Dave$^{1}$, Peter Hofer$^{2}$, Jan Forstner$^{3}$,
Marlene Villneuve$^{3}$,\\ Robert Galler$^{3}$, and Elmar Rueckert$^{1}$
\thanks{$^{1}$Chair of Cyber-Physical System, Montanuniversität Leoben, Austria.}%
\thanks{$^{2}$Theresianische Militarakademie, Austria.}%
\thanks{$^{3}$Chair of Subsurface Engineering, Montanuniversität Leoben, Austria.}%
\thanks{Corresponding Author: \tt\small linus.nwankwo@unileoben.ac.at}
}

\begin{document}

\maketitle
\thispagestyle{empty}
\pagestyle{empty}

\begin{abstract}
To ensure the efficiency of robot autonomy under diverse real-world conditions, a high-quality heterogeneous dataset is essential to benchmark the operating algorithms' performance and robustness. Current benchmarks predominantly focus on urban terrains, specifically for on-road autonomous driving, leaving multi-degraded, densely vegetated, dynamic and feature-sparse environments, such as underground tunnels,  natural fields, and modern indoor spaces underrepresented.
To fill this gap, we introduce EnvoDat, a large-scale, multi-modal dataset collected in diverse environments and conditions, including high illumination, fog, rain, and zero visibility at different times of the day. Overall, EnvoDat contains 26 sequences from 13 scenes, 10 sensing modalities, over 1.9TB of data, and over 89K fine-grained polygon-based annotations for more than 82 object and terrain classes. We post-processed EnvoDat in different formats that support benchmarking SLAM and supervised learning algorithms, and fine-tuning multimodal vision models. With EnvoDat, we contribute to environment-resilient robotic autonomy in areas where the conditions are extremely challenging. The datasets and other relevant resources can be accessed through \url{https://linusnep.github.io/EnvoDat/}.
\end{abstract}

\section{INTRODUCTION}
Humans can accurately build a mental map of an environment (whether known or unknown), describe their location, and that of objects in the environment, and return to their initial position effortlessly. However, adapting autonomous agents to perform such innate abilities and operate reliably across diverse environments demands highly accurate and robust geospatial perception and simultaneous localisation and mapping (SLAM) algorithms~\cite{Stachniss2016, slam1}.
\begin{figure}[htp]
 \centering
  \includegraphics[scale=0.24]{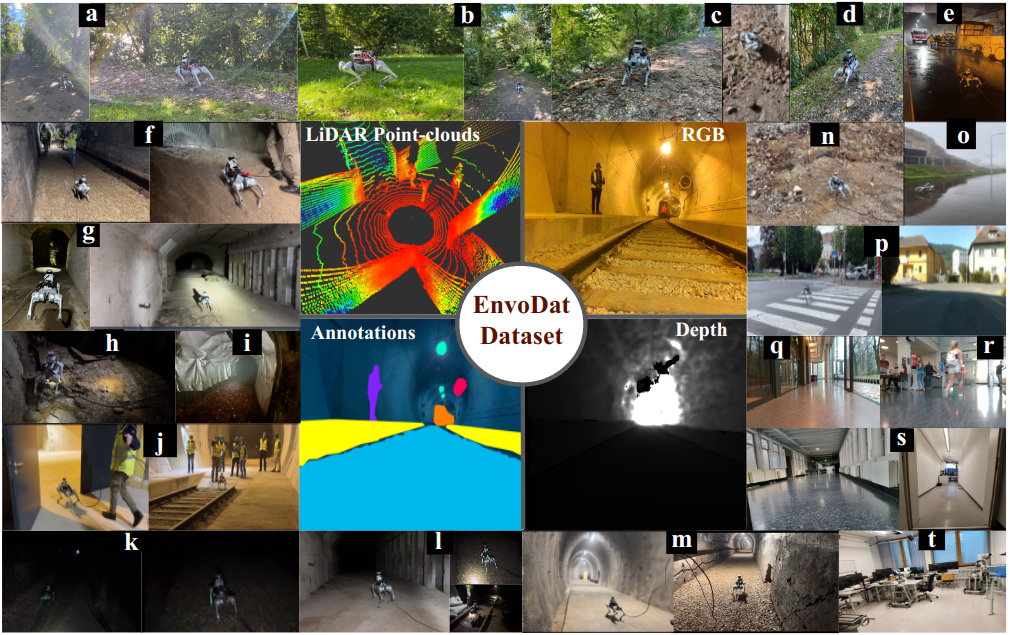}
 \caption{An overview of scene diversity in EnvoDat. EnvoDat includes time-synchronized multimodal sensor data (e.g., RGB, LiDAR, depth) and fine-grained annotations, captured in challenging mixed in-outdoor, off-road, and subterranean environments under several real-world environmental conditions such as poor illumination, zero visibility, foggy, smoky, etc.}
\label{fig:EnvoDat}
\vspace*{-1.25\baselineskip}
\end{figure}
Despite the effort made so far by the robotics community to develop scalable and generalisable ecocentric robotic perception and SLAM algorithms, many of the state-of-the-art (SOTA) algorithms still struggle in real-world deployment \cite{Nwankwo_2023, kaveti2023challenges}. This is particularly common in multi-degraded, densely vegetated, dynamic and feature-sparse heterogeneous environments, such as underground tunnels, natural fields, public spaces, and modern indoor spaces with specular surfaces \cite{Zhao2024CVPR, 10415477}.

Common benchmarks such as KITTI \cite{kitti}, Oxford RobotCar \cite{roboCar} and TUM-VIE \cite{tumVi}, which were collected in on-road urban and controlled indoor environments, are inadequate for these complex off-road environments. Algorithms benchmarked on these datasets often fail to generalise to off-road or subterranean terrains due to the inherent discrepancies in the environments' features and geometric characteristics.
Furthermore, real-world environments are often in a state of constant flux due to maintenance, constructions, or being abandoned (see Fig.\ref{fig:EnvoDat}).  This viability poses challenges for accurate perception and SLAM in autonomous agents.
Consequently, relying on datasets collected long ago as benchmarks for contemporary perception and SLAM algorithms can potentially lead to inaccuracies.  This is because, they do not account for spatio-temporal changes, and advancement in sensor specifications and modalities. Therefore, frequent data collection to capture the intricacies or temporal change in such environments stands paramount. This forms one of our core motivations for developing the EnvoDat dataset. 
Specifically, our contributions are fourfold:
\begin{itemize}
    \item We introduce the EnvoDat dataset which encompasses 10 rich sets of sensory modalities, $>$~1.9TB of data, and over 89K fine-grained, layered polygon-based human annotated RGB images suitable for object detection, classification, and segmentation tasks.
    \item We captured both temporal and structural diversities of multi-heterogeneous environments, and we provided insights into how the heterogeneity conditions affect the spatial perception and reasoning in those environments. 
    \item We performed a benchmark evaluation of the SOTA approaches on our dataset to examine how real-world environmental conditions e.g., dynamic entities, visibility, etc, affect the performance of the algorithms.
    \item We made our dataset public, including post-processing tools to accelerate the development of environment-invariant perception and SLAM algorithms.
\end{itemize}

\section{Prior works}\label{B} 
Several datasets have been introduced over the years~\cite{9968057, hilti, kitti}, and each has its specific environmental focus and conditions. We reviewed the key attributes of these datasets, e.g., the environments in which they were collected, the sensory modalities they employed, and the specific conditions under which they were curated. We compared them against the EnvoDat dataset in Table \ref{tab:literature}.

\begin{table*}
    \caption{Review of common SLAM datasets, and comparison with the EnvoDat dataset}
    \vspace{-7.5pt}
    \centering
    \begin{tabular}{c|c c c c c c|c c c c|cccccc}
    \hline
        \multirow{2}{*}{\textbf{Datasets}} & \multicolumn{6}{c|}{\textbf{Environments \& Focus}} & \multicolumn{4}{c|}{\textbf{Sensory Modalities \& Platforms}} & \multicolumn{6}{c}{\textbf{Collection Conditions}}\\
         & I & U & S & V & SSV & Div & LiD & Cam & IMU & Motion & Dyn & ANC & AgM & LC & Vis & RT\\
    \hline
        TUM-RGBD \cite{tum} & $\checkmark$ & $\times$ & $\times$ & $\times$ & low & low & $\times$ & $\checkmark$ & $\times$ & HH,WR & $\checkmark$ & $\times$ & $\checkmark$ & $\times$ & FV & $\times$\\
        RoboCar \cite{roboCar} & $\times$ & $\checkmark$ & $\times$ & $\times$ & low & low & $\times$ & $\checkmark$ & $\checkmark$ & Veh &$\checkmark$ & $\checkmark$ & $\times$ & $\times$ & FV & $\times$\\
        Wild-Places \cite{wild-places} & $\times$ & $\checkmark$ & $\times$ & $\checkmark$ & low &low & $\checkmark$ & $\times$ & $\times$ & HH & $\times$ & $\times$ & $\times$ & $\times$ & FV & $\times$\\
        KITTI \cite{kitti} & $\times$ & $\checkmark$ & $\times$ & $\times$ & low & low & $\checkmark$ & $\checkmark$ & $\times$ & Veh & $\checkmark$ & $\times$ & $\checkmark$ & $\times$ & FV & $\times$\\
        SUN-RGBD \cite{sun} & $\checkmark$ & $\times$ & $\times$&  $\times$&low &low & $\times$ & $\checkmark$ & $\times$ & HH & $\times$ & $\times$ &  $\times$& $\times$ & FV & $\times$\\
        Cityscapes \cite{7780719} &$\times$&$\checkmark$&$\times$&$\times$&low& high &$\times$ & $\checkmark$ & $\times$ & Veh & $\checkmark$ & $\times$ & $\times$ & $\times$ & FV & $\times$\\
        TUM-VIE \cite{tumVi} & $\checkmark$ & $\checkmark$ & $\times$ & $\times$ &low&low & $\times$ & $\checkmark$ & $\checkmark$ & HH,HM & $\checkmark$ & $\times$ & $\checkmark$ & $\checkmark$ & FV & $\times$\\
        NAVER \cite{locdata} & $\checkmark$ & $\times$ & $\times$ & $\times$ & low& low& $\checkmark$ & $\checkmark$ & $\times$ & WR & $\checkmark$ & $\times$ & $\times$ & $\checkmark$ & FV & $\checkmark$ \\
        Comp.Urban \cite{complexUrban} &$\times$& $\checkmark$ & $\times$& $\times$ & medium & low & $\checkmark$ & $\checkmark$ & $\checkmark$ & Veh &$\checkmark$ & $\checkmark$ & $\times$ & $\times$ &FV & $\times$\\
        CADC \cite{cadc} & $\times$ & $\checkmark$ & $\times$ & $\times$ & low & low  & $\checkmark$ & $\checkmark$& $\checkmark$& Veh &$\checkmark$ & $\checkmark$ & $\times$ & $\times$  & FV, PV & $\times$ \\
        Hilti-Oxford \cite{9968057} & $\checkmark$ & $\times$  & $\times$& $\times$ &medium & low&  $\checkmark$& $\checkmark$ &$\checkmark$ &HH &$\times$ & $\times$& $\checkmark$& $\times$&FV&$\times$\\ 
        TAIL \cite{10542164} & $\times$ & $\checkmark$ & $\checkmark$& $\times$ & low& low& $\checkmark$ & $\checkmark$ &$\checkmark$ & WR, Quad& $\times$& $\times$& $\checkmark$ &$\times$&FV&$\times$\\ 
        SubT-MRS \cite{Zhao2024CVPR} & $\checkmark$ & $\checkmark$ & $\checkmark$ & $\times$ & high & high & $\checkmark$ & $\checkmark$ & $\checkmark$& Multi&$\checkmark$ &$\checkmark$ &$\checkmark$  &$\checkmark$ &ALL&$\times$\\ 
        MulRan \cite{9197298} & $\times$ & $\checkmark$ &$\times$ & $\times$ & low& low& $\checkmark$ & $\times$ & $\times$& Veh&$\checkmark$ & $\times$& $\times$ &$\times$ &FV & $\times$\\
        CODa \cite{10530418} & $\times$ &  $\checkmark$& $\times$&  $\times$& low& low& $\checkmark$ & $\checkmark$ & $\checkmark$& WR& $\checkmark$&$\times$ &$\times$  &$\times$&FV&$\times$\\
        \hline
       \textbf{EnvoDat (ours)} & \textbf{$\checkmark$} & \textbf{$\checkmark$} & \textbf{$\checkmark$} & \textbf{$\checkmark$} & \textbf{high} & \textbf{high} & \textbf{$\checkmark$} & \textbf{$\checkmark$} & \textbf{$\checkmark$}& \textbf{Quad, WR} & \textbf{$\checkmark$} & \textbf{$\checkmark$} &\textbf{$\checkmark$} & \textbf{$\checkmark$} & \textbf{ALL} &\textbf{$\checkmark$}\\
    \hline
    \vspace{-5pt}
    \end{tabular}
    *\textbf{Legends:}
   \textbf{Environments \& Focus} (I $\rightarrow$ Indoor, U $\rightarrow$ Urban, S $\rightarrow$ Subterranean and granular e.g., tunnels and desert scenes. V $\rightarrow$ densely vegetated area, e.g., forest. SSV $\rightarrow$ Spatial scale variation of the environment. Div $\rightarrow$ Scene diversity). \textbf{Sensory Modalities \& Platforms} (LiD $\rightarrow$ 3D LiDAR, Cam $\rightarrow$ RGB-D cameras, IMU $\rightarrow$ Inertial measurement units; Motion: HH $\rightarrow$ Handheld, WR $\rightarrow$ Wheeled robot, Veh $\rightarrow$ Vehicle, HM $\rightarrow$ Head mount, Quad $\rightarrow$ Quadruped robot). \textbf{Collection Conditions} (Dyn $\rightarrow$ Several dynamic objects, ANC $\rightarrow$ Adverse and natural conditions e.g., fog, rain, smoky, wet, rocky, sandy and soft terrains. AgM $\rightarrow$ Aggressive motions e.g., jerky, fast, sudden turns, gradient descending and ascending (unstructured terrains), and sudden changes in direction. LC $\rightarrow$ lighting conditions e.g., varying illumination, day or artificial lightening, Vis $\rightarrow$ Visibility conditions (ALL: FV $\rightarrow$ Fully visible, PV $\rightarrow$ Partially visible, and ZV $\rightarrow$ Zero visibility), RT $\rightarrow$ Specular or reflective and transparent surfaces).
\label{tab:literature}
\end{table*}

\subsection{Urban, Indoor, and Natural Environment Datasets}\label{B1}
The datasets presented in Table \ref{tab:literature} are some of the most popular existing benchmarks. The majority of them focus on urban on-road environments, particularly for the outdoor autonomous drive (e.g., KITTI \cite{kitti}, Oxford RobCar \cite{roboCar}, CADC \cite{cadc}, etc.). Others focus on urban place recognition, e.g., MulRan \cite{9197298}, HeliPR \cite{doi:10.1177/02783649241242136}, CODa \cite{10530418}, Cityscapes \cite{7780719}, Warburg et al. \cite{Warburg_2020_CVPR}, etc.
For indoor environments, datasets that capture long-term trajectories with loop closure and challenging factors like surface reflectivity, transparency, and dynamics \cite{Nwankwo_2023}, which are critical for sensors like LiDAR and RGB-D cameras are lagging. The TUM benchmarks \cite{tum}, \cite{tumVi}, ICL \cite{Characterizing19}, VECtor \cite{9809788}, SUN RGB-D \cite{sun}, and SceneNet RGB-D \cite{McCormac_2017_ICCV} often focus on small spatial (e.g., single room), fully observable, feature-dense environments with few or no dynamic objects.
In Table \ref{tab:literature}, we refer to these focus as spatial scale variation (SSV) and scene diversity (Div). High SSV and Div for indoor (I) and urban (U) environments include datasets with a wide range of spaces, from small rooms to large halls, and narrow alleys to wide streets. Low SSV and Div focus on uniform spaces, like only small rooms or similar-sized urban areas. For subterranean (S) and vegetated (V) environments, high SSV and Div cover varied spaces, such as narrow tunnels to large caverns, and different vegetation types from dense forests to open fields. Low SSV and Div cover similar-sized tunnels or specific vegetation types without much variation.

Furthermore, existing urban and indoor datasets like SUN RGB-D \cite{sun}, TUM benchmarks \cite{tum, tumVi}, and Yin et al.\cite{Yin2023GroundChallengeAM} focus solely on vision-based scene understanding with RGB-D cameras. These datasets are not optimal for benchmarking perception or SLAM algorithms intended for scenarios such as dawn search and rescue or late-night inspection tasks.

Similar to urban and indoor environments, efforts have been made to capture data in natural and multi-degraded terrains. Recent datasets like Wild-places \cite{wild-places}, SubT-MRS \cite{Zhao2024CVPR}, CitrusFarm dataset \cite{citrusFarm}, BotanicGarden \cite{10415477}, and Tail \cite{10542164} address challenges posed by off-road and natural settings, but offered limited scope in terrain variability and diverse geometric and feature characteristics. Generally, off-road and natural terrain datasets, especially underground tunnels, are under-represented. 
Therefore, there's a need for a holistic dataset that encompasses a wide range of diverse environments. EnvoDat fills this crucial gap.

\section{EnvoDat dataset - The overview}\label{sec:3}
We recorded the EnvoDat dataset in challenging real-world in-outdoor and subterranean environments of different geometric and feature characteristics, depicting realistic operations scenarios for autonomous robots. 
The data diversity and exemplary features and geometric characteristics of the scenes are shown in Figure \ref{fig:EnvoDat}. We captured multiple sequences in 13 scenes with over 1.1M inertial measurement samples at \(100 Hz\), more than 600K 3D LiDAR data including 2D scan and four image data layers from a long-range 32-channel Ouster LiDAR (up to 45m) at \(10 Hz\). We also captured over 500K RGB and depth (RGB+D) images of an Intel Realsense D435i sensor at \(30\) frame per second (fps), and over 100K RGB images of ELP 4K monocular camera at \(30\) fps. 
\begin{table*}[htp]
\caption{Core statistics of the EnvoDat dataset. Refer to \cite{envodat} for details of the individual scene statistics}
\vspace{-7.5pt}
  \centering
\begin{tabular}{c c c c c c c c c c c  c c c}
\toprule
 \textbf{Sc.Typ} & \textbf{NSeq} &
  \textbf{\parbox{1cm}{\centering{LiDAR\\ PtC}}} & \textbf{\parbox{1cm}{\centering{LiDAR\\Img lay}}} & \textbf{2D Scan} & \textbf{RGB+D} & \textbf{ImS} &  \textbf{Mono} & \textbf{CT} & \textbf{Size} & \textbf{Dur} & \textbf{NIm} & \textbf{NAn} &\textbf{OTC}\\
 \midrule
                                Indoor & 12 &46.7K+ & 187.4K+ &46.7K+& 260.7K+ & 468.5K+ & 111.3K+ & M,A,E & 798.6 & 1.34 &2.9K+&34.8K+&35\\
                             
                                Outdoor & 3 &21.1K+& 85.5K+ & 21.1K+& 115.6K+& 210.2K+ & - &M,A&486.5&0.59&4.7K+&45.9K+&36\\
      
                                SubT& 11 &47.7K+&190.8K+&47.7K+& 141.2K+ &471.5K+& - &M,A&698.7&1.93&0.9K+&8.6K+&46\\
\bottomrule
\vspace{-5pt}
\end{tabular}
*\textbf{Legends:} \textbf{Sc.Typ} $\rightarrow$ Scene type, details of the individual scene, can be found at \cite{envodat}.  \textbf{SubT} $\rightarrow$ Underground tunnels. \textbf{NSeq} $\rightarrow$ Number of sequence. \textbf{LiDAR PtC} $\rightarrow$ 3D LiDAR point clouds data in pcd format at $10Hz$. \textbf{LiDAR Img lay} $\rightarrow$ Four 2D image data layers of the 3D LiDAR (the signal, reflectivity, range, and near-infra-red image frames) at $10Hz$. \textbf{RGB+D} $\rightarrow$ RGB and depth image frames of Intel Realsense D435i camera at $30fps$. \textbf{ImS} $\rightarrow$ IMU data samples at $100Hz$. \textbf{Mono} $\rightarrow$ RGB images of ELP 4K monocular camera at $30fps$. \textbf{CT} $\rightarrow$ Data collection time, M = morning, A = Afternoon, and E = Evening. \textbf{Size} $\rightarrow$ Estimated download size including the uncompressed ROS bag files in gigabytes (GB).
\textbf{Dur} $\rightarrow$ Duration of raw data recording in hours. \textbf{NIm} $\rightarrow$ Number of annotated image frames. \textbf{NAn} $\rightarrow$ Number of annotations. \textbf{OTC} $\rightarrow$ Number of object and terrain classes.
\label{tab:statistics}
\vspace*{-1.75\baselineskip}
\end{table*}
Further, we provide over 89K human annotations of RGB images for 82 objects and terrain classes (e.g., infrastructure and transport: rail track, car, train, excavator; safety: fire extinguisher, exit sign, helmet;  hazard: debris, water pool, rust iron; nature and vegetation: grass, tree, sky, hill, fog; human: person, cyclist etc). For the complete list, and semantic category mapping, refer to \cite{envodat} and Fig.\ref{fig:annotatation}.
The core data statistics are summarised in Table \ref{tab:statistics}.

\subsection{Selection of Environments and Sequences}\label{sec:3A}
We selected each scene based on distinct feature attributes: 
(i) Leo-For (Fig.\ref{fig:EnvoDat}a-d) is a typical-vegetated area with unstructured terrain (i.e., sloppy, elevated), dense canopies, varied light penetrations, and dynamic shadow formations caused by the swaying of leaves and branches. These features present distinct challenges to robotic perception and spatial awareness. 
(ii) Leo-Str (Fig.\ref{fig:EnvoDat}p) is an open street in Leoben, Austria, characterised by sparse features, pedestrians, vehicles, and several buildings. These ranges of environmental complexities are often not constant, and poses substantial challenges to perception and SLAM systems used in inspection, monitoring and surveillance in such environments.

(iii) SubT-ZAB and SubT-SBe (Fig.\ref{fig:EnvoDat}e-m) are highly symmetric and under-maintenance sub-tunnels, Zentrum Am Berg (ZAB) and Schloss Berg (SBe) located in Eisenerz and Graz, Austria. These scenes feature a variety of visually and geometrically degraded conditions, including poor illumination, partial visibility, inconsistent brightness, sparse features, textureless and rust surfaces, mixed of narrow to large symmetry, etc.  These challenging characteristics complicate feature matching and can lead to errors in pose estimation, collision risks, and inaccuracies in map reconstructions.
Additionally, the absence of GPS, within these underground tunnels, makes accurate localization difficult, which is crucial for autonomous navigation and mapping tasks \cite{articleGPS}.
(iv) MU-Hall (Fig.\ref{fig:EnvoDat}r) is a mid-large exhibition hall. The first sequence (MU-Hall-seq-01) was recorded during an exhibition with many dynamic and static features (i.e., moving people, and several pieces of exhibition furniture).
(v) MU-Cor (Fig.\ref{fig:EnvoDat}s) is a high symmetric passageway of approximately $6$ x $120\;m$ with several tables and natural lighting conditions.
(vi) MU-CPS (Fig.\ref{fig:EnvoDat}t) depicts a typical office layout with multiple occlusions, several pieces of furniture, different lighting conditions (i.e., natural daylight and artificial lights), and multiple loop closures. (vii) the MU-TXN (Fig.\ref{fig:EnvoDat}q) is the student study centre at the Montanuniversität characterised by large transparent and reflecting surfaces, and numerous moving objects, creating a mixed static and dynamic environment.
Each of the aforementioned attributes can severely degrade robotic autonomy and spatial awareness in heterogeneous environments. For detailed information about the scene characteristics, refer to the \textbf{\textit{Scenes}} section of  \cite{envodat}. 

\subsection{Sensors Setup, Synchronisation and Calibration}
In each scene of the EnvoDat, we captured raw sensor data using different sensor suites. The 
 sensor specifications, setup and calibration are described in \cite{envodat}. 
Each sensor in the suite transmits data at its own rate and with individual timestamps, which can cause slight timing variations. 
To address this, we employed the approximate time policy of the ROS \cite{ros} message filters framework to synchronise the timestamps of the incoming data across different sensory modalities. This approach ensured a few seconds of accuracy and maintained time-synchronisation and the overall consistency of the fussed sensor data across all the scenes. 

\subsection{Data Acquisition and Robot Platforms}\label{D2}
We intermittently recorded the dataset between November 2022 and September 2024, using two robot platforms, due to the nature of the environments. For indoor environments with a structured layout, we utilised our open-source wheeled robot~\cite{nwankwo2023romr}. However, for off-road and outdoor environments with unstructured terrains,  we utilised our Unitree Go1 quadruped robot. We captured the raw data by teleoperating the robots around the environments.
Leveraging the ROS framework \cite{ros}, we recorded the following data: 

For the LiDAR and IMU sensors, we captured inertial data, point clouds, laser scans, and four 2D image data layers: signal, reflectivity, range, and near-infra-red (NIR) image data.
The signal images represent the strength of the light returned to the LiDAR sensor measured in the number of photons of light detected. The reflectivity images capture the reflectance of surfaces detected by the sensor. Range images measure the distance of a point from the sensor, based on the time of flight of the laser pulse. NIR images show the sunlight intensity at the \(865 nm\) wavelength, measured in the number of photons not generated by the sensor’s laser pulse.

For the vision-based sensors, we captured the camera info, RGB and depth images, all recorded at 30 frames per second (fps). The RGB and depth resolution are $1920$ x $1080$ and $1280$ x $720$ respectively. All the data from the scenes were initially collected in ROS bag format and post-processed in other formats suitable for benchmarking SLAM, perception and supervised learning algorithms.

\subsection{Ground-truth Poses}\label{3C}
\begin{figure}[htp]
 \centering
 \vspace*{-1.25\baselineskip}
 \subfigure[Geospatial map]{\includegraphics[scale=0.104]{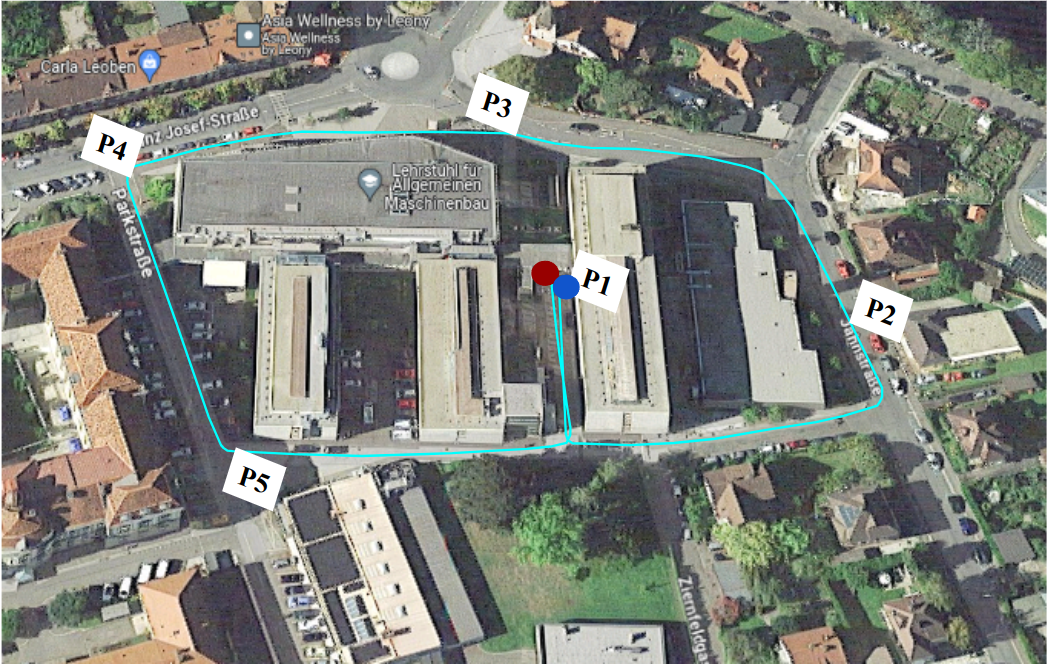}}
 \subfigure[Pseudo-ground truth map]{\includegraphics[scale=0.11]{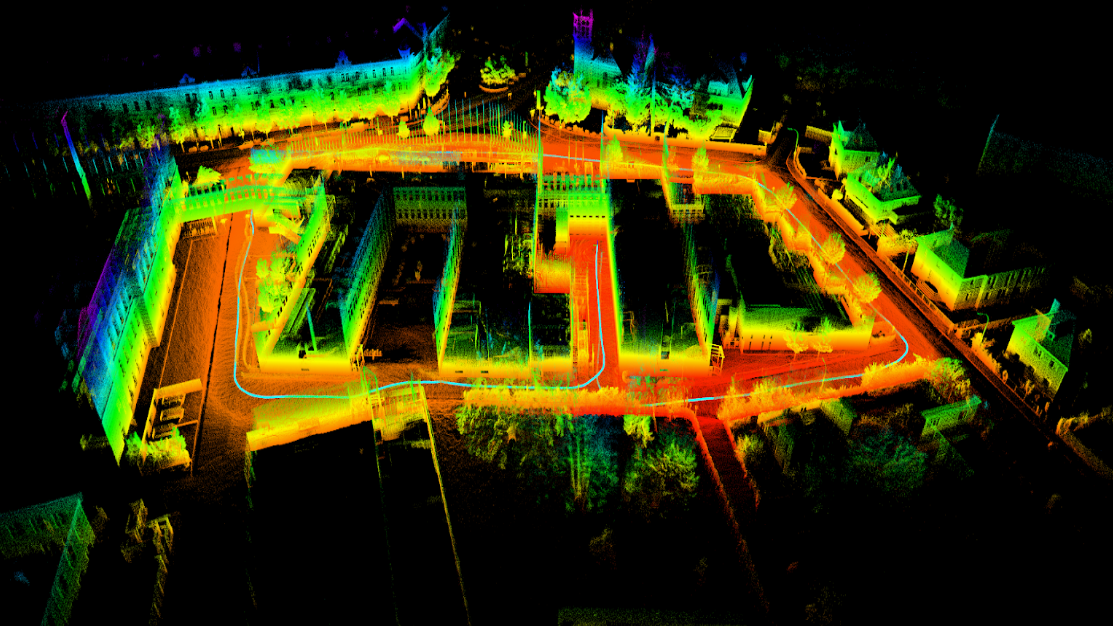}}
 \subfigure[Spatial reconstruction]{\includegraphics[scale=0.27]{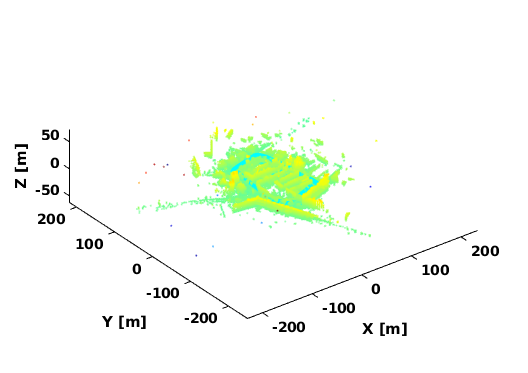}}
  \subfigure[Features distribution]{\includegraphics[scale=0.23]{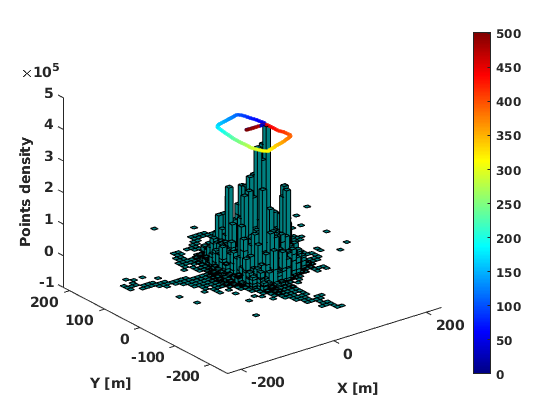}}
 \caption{Geographic map and the reconstructed maps in one of the scenes in EnvoDat. (a) Satellite image with annotated trajectory and waypoints \textbf{P1 - P5}, where we slowed the robot for multiple measurements to refine pose estimates. (b) Dense LiDAR inertial odometry pseudo-ground truth reconstruction. (c) Reconstructed map from raw LiDAR point clouds for scale precision. (d) Feature point distribution from the reconstructed map, showing high-density regions where the robot slowed and low-density areas of faster movement. The robot's trajectory, overlaid on the density bars, is colour-coded based on the cumulative distance transitioning from blue (start) to red (end). This data supports the correlative analysis of the impact of feature densities on SLAM algorithms performance in Subsection \ref{featDensities}.}
\label{fig:reconMaps}
\vspace*{-1.05\baselineskip}
\end{figure}

We provide high-precision reference ground-truth data for each scene, tailored to the scene's features and geometric characteristics. Since most of the scenes in the EnvoDat such as the indoor and the sub tunnels are GPS denied, we obtained the pseudo-ground truth data from the fusion of different constraints derived from LiDAR inertial estimations from GLIM~\cite{glim}. To verify the robustness and accuracy of these pseudo-ground truth trajectories, we compared the reconstructed LiDAR inertial odometry map with the ground truth geospatial map of the scene. Additionally, we assessed the scale precision by spatially reconstructing the map of the scene using the raw LiDAR sensor's point cloud data.
Figure \ref{fig:reconMaps}(a-c) shows our quantitative and qualitative assessment, obtained by overlaying the global estimated trajectory poses on both the geospatial and the 3D spatial reconstructed maps.

\subsection{Data Annotations and Ontology}\label{3F}
\begin{figure}[htp]
 \centering
 \vspace*{-1.0\baselineskip}
 \subfigure[Fine-grained annotation]{\includegraphics[scale=0.1163]{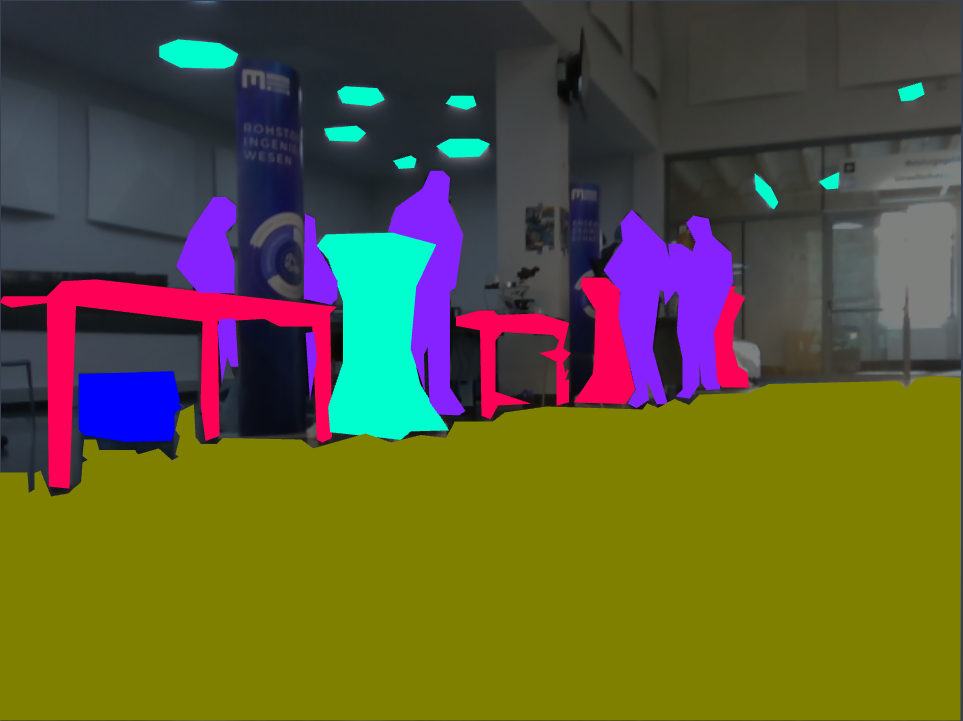}}
 \subfigure[Annotations and categories]{
 \includegraphics[scale=0.195]{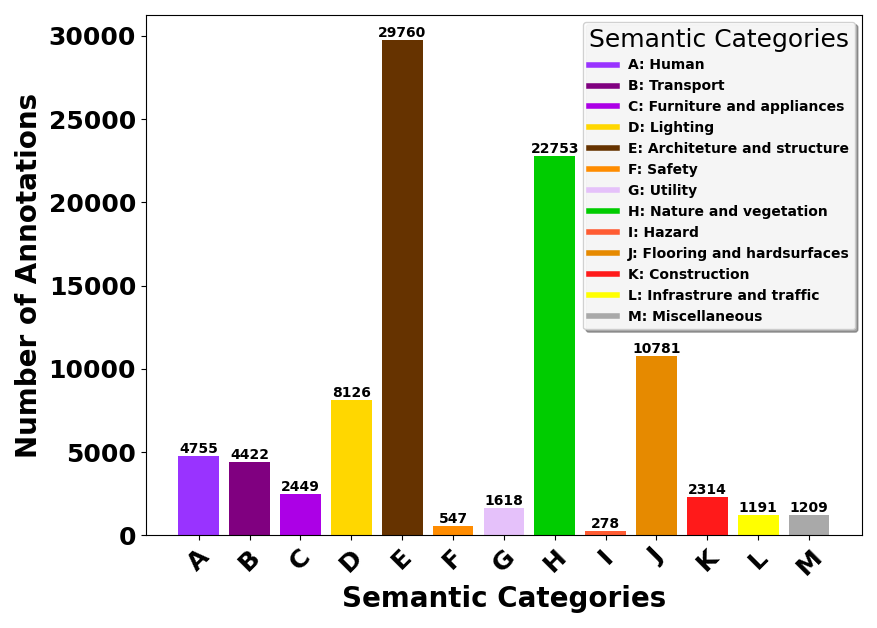}}
 \caption{A subset of the EnvoDat annotation. We provided over 89K annotations, with an average of 10.3 annotations per image across 82 object and terrain classes. 
 We provide the annotated data in different formats e.g., JSON, XML, TXT, and CSV to support fine-tuning vision models.}
\label{fig:annotatation}
\vspace*{-0.5\baselineskip}
\end{figure}

We prepared our dataset to support various applications including, (a) supervised learning algorithms, and (b) language and visual foundation models. 
From the extracted RGB images across the 13 scenes in the EnvoDat dataset, we defined objects and terrain classes visible in the camera's field of view.
We provide three weeks of frame-wise annotations, 3 hours per day for over 89K RGB annotations distributed across the scenes using Roboflow \cite{roboflow}. Currently, EnvoDat includes 82 objects and terrain classes, and we aim to scale these numbers to include more objects, terrain classes, and temporal variations in future updates.
To ensure high-quality annotations and minimize class overlaps, we utilized fine-grained, layered polygon-based annotations instead of the traditional bounding boxes (see Fig.\ref{fig:annotatation}(a)). 
Given the scene diversities in the EnvoDat, we manually selected frames that highlight key features, geometric details, and environmental challenges such as dynamic objects, dense vegetation, opaque surfaces, and varied terrains for annotation. Each object and terrain class are mapped to their corresponding semantic categories, as shown in Fig.\ref{fig:annotatation}b.

In the post-annotation stage, we conducted a manual quality check on the annotated data to ensure that at least $90\%$ of the annotations are valid and accurately correspond to the intended object and terrain classes. For more details of our annotation ontology and labelling policy, see the \textbf{\textit{Annotations}} section of the project site \cite{envodat}.

\section{Experiments}
In our experiments, we investigate how real-world conditions can affect robot autonomy in heterogeneous environments. We deployed the EnvoDat dataset for four baseline applications - mapping, localisation, object detection, and classification. We designed three experiments that address the following research questions (RQs):
\begin{itemize}
     \item RQ1 - Is the performance of the SOTA SLAM algorithms significantly degraded by environment-specific conditions e.g., dynamic entities, varying illumination, opaque surfaces, partial visibility conditions, etc.?
    \item RQ2 - To what extent does the feature density or sparsity influence robotic autonomy and scene perception in heterogeneous environments?
    \item RQ3 - How do the heterogeneity of real-world environments and the viability of objects and terrain appearances, lighting conditions, non-standard objects, etc observed in the majority of the scenes in the EnvoDat affect the object detector models trained on common household, urban or controlled environment datasets?
\end{itemize}

\subsection{SLAM Algorithms Benchmark}\label{sec:4A}
To address RQ1, we benchmarked five SOTA SLAM algorithms on our dataset. We selected two visual-based methods (RTAB~\cite{rtab} and ORB-SLAM3~\cite{orbslam}), one graph-based LiDAR method (HDL-SLAM~\cite{hdl}), and two filter-based LiDAR methods (FAST-LIO2~\cite{xu2022fast} and DLIO~\cite{chen2022dlio}).
We chose five scenes from EnvoDat, which exhibits at least some of the geometric and feature characteristics highlighted in RQ1: (i) dynamic entities - MU-Hall-01, (ii) varying illumination and opaque surfaces - MU-TXN-01, (iii) zero and partial visibility conditions - SubT-ZAB-01 and MU-Cor-03 (night sequence), and (iv) high-dimensional observations \& features-sparsity - Leo-Str-01.
We analysed the impact of these environmental factors on the mapping accuracy by comparing the mapped points and estimated trajectories to the ground truth data (see Table \ref{tab: indoorstat}).
We defined per-point absolute trajectory error (ATE), relative pose error (RPE) and scale drift (SD) as evaluation criteria.

Formally, given a sequence of poses from the estimated trajectory \(\hat{\mathbf{T}} \triangleq [\hat{\mathbf{T}}^{tx}, \hat{\mathbf{T}}^{ro}]^{\top} = [\hat{\mathbf{t}}_1, \hat{\mathbf{t}}_2, \ldots, \hat{\mathbf{t}}_n]^\top \text{, \>} \hat{\mathbf{t}}_i \in \mathbb{R}^3\) and the ground truth trajectory \(\mathbf{T} \triangleq [\mathbf{T}^{tx}, \mathbf{T}^{ro}]^{\top} = [\mathbf{t}_1, \mathbf{t}_2, \ldots, \mathbf{t}_n]^\top, \> \mathbf{t}_i \in \mathbb{R}^3\) at the \(i-th\) timestamp, we compute per point ATE (ppATE) as follows. The superscripts \(tx\) and \(ro\) denote the translational and rotational components of the homogeneous transformation matrix of \(\hat{\mathbf{T}}\) and \(\mathbf{T}\). In all our evaluations, we utilised only the translational components.
  \( \text{ppATE}_i = \|\hat{\mathbf{t}}_i - \mathbf{t}_i\|^2 \quad \forall i \in \{1, 2, \ldots, n\} \)
where  $||.||$ is the Euclidean norm of the position difference between the \(\hat{\mathbf{T}}\) and \(\mathbf{T}\) poses. The overall ATE is therefore computed as depicted in Eq.\ref{eqnATE}.
\begin{equation}\label{eqnATE}
    \text{ATE} = \left( \frac{1}{n} \sum_{i=1}^n \text{ppATE}_i \right)^{1/2} = \sqrt{\frac{1}{n} \sum_{i=1}^n \|\hat{\mathbf{t}}_i - \mathbf{t}_i\|^2}
\end{equation}

Unlike the ATE, which measures the global consistency of the entire trajectory, we utilized the RPE to independently estimate drift between poses over fixed-length segments of the trajectory.
Formally, let \(\Delta\) be the fixed-length segments of the trajectory. For each segment \(i\), we compute the per-point RPE as 
$\text{ppRPE}_{i-\Delta} = \|\hat{\mathbf{r}}_i - \mathbf{r}_i\|^2 \quad \forall i \in \{2, 3, \ldots, n\}$, where $ \hat{\mathbf{r}}_i = \hat{\mathbf{t}}_i - \hat{\mathbf{t}}_{i-\Delta} \quad \text{and} \quad \mathbf{r}_i = \mathbf{t}_i - \mathbf{t}_{i-\Delta}$.
The overall RPE is therefore computed as the mean of the individual errors over all segments, as described in Eq.\ref{eqnRPE_trans}.
\begin{equation}\label{eqnRPE_trans}
\text{RPE} = \frac{1}{n-\Delta} \sum_{i=1}^{n-\Delta} \text{ppRPE}_{i-\Delta} = \frac{1}{n-\Delta} \sum_{i=1}^{n-\Delta} \|\hat{\mathbf{r}}_i - \mathbf{r}_i\|^2
\end{equation}
Eq. \ref{eqnRPE_trans} is highly sensitive to the choice of \(\Delta\). A small \(\Delta\) will measure the local consistency between closely spaced poses, while a large \(\Delta\) will evaluate consistency over longer trajectory segments. For all our evaluations, we set \(\Delta = 1\).

Furthermore, given the sequence of poses from the two trajectories \(\hat{\mathbf{T}}^{tx}\) and \(\mathbf{T}^{tx}\), we computed the scale drift \( SD \) by comparing their cumulative distances over the \(\Delta\) i.e.,
\begin{equation}\label{eqnSD}
SD = \frac{1}{n - \Delta} \sum_{i=1}^{n-\Delta} \left| \frac{\lVert \mathbf{\hat{t}}_i - \mathbf{\hat{t}}_{i+\Delta} \rVert}{\lVert \mathbf{t}_i - \mathbf{t}_{i+\Delta} \rVert} - 1 \right|
\end{equation}
From Eq. \ref{eqnSD}, if the algorithm performed optimally (i.e., with no deviation), then SD = 1. SD values less than or greater than 1 indicate underestimation or overestimation. We used SD to evaluate the robustness and stability of the algorithm's scale estimation over different distances and time frames.

Table \ref{tab: indoorstat} shows the benchmark results of the SLAM algorithms, with the best metrics in bold. Most of the algorithms struggled with the complexities of the environments. However, regardless of the environments' features and geometric characteristics, Fast-LIO2~\cite{xu2022fast} and DLIO~\cite{chen2022dlio} consistently outperformed the other algorithms across all the scenes, with comparatively lower ATE and RPE values.
Conversely, the visual-based methods failed in the subterranean and night sequences due to poor visibility conditions. The HDL-SLAM~\cite{hdl} also showed competitive results in some of the scenes; however, it does not match the overall robustness of Fast-LIO2~\cite{xu2022fast}, which stands out as the top-performing algorithm in the benchmark.
\begin{table*}[h]
\caption{Results from SLAM algorithms benchmark. $(S_{n})$ is the sequence number, $n = 1 -(\text{Morning}), 2 - \text{Afternoon}, 3 - \text{Eveneing}$}
\vspace{-7.5pt}
  \centering
\begin{tabular}{c| c|c|c|c|c|c}
\hline
\multirow{2}{*}{\textbf{Methods}} & \multirow{2}{*}{\textbf{Metrics}} & \multicolumn{1}{c|}{\textbf{Dynamic Entities}} & \multicolumn{1}{c|}{\textbf{V.Illum. \& Opaque}} & \multicolumn{2}{c|}{\textbf{Zero \& Partial Visibility}} & \multicolumn{1}{c}{\textbf{HD Obs. \& F-Spar}}\\
\cline{3-7}
 && \textbf{MU-Hall $(S_{1})$} & \textbf{MU-TXN $(S_1)$} & \textbf{SubT-ZAB-01 $(S_2)$} & \textbf{MU-Cor $(S_3)$} & \textbf{Leo-Str $(S_1)$}\\
 \hline
 \multirow{3}{*}{HDL-SLAM~\cite{hdl}}
                                & ATE ($\mu / \sigma$) & 17.78 / 7.76 & 45.86 / 21.44 & 7.54 / 4.90 & 5.12 / 3.83 & 3.29 / 6.60\\
                                & RPE ($\mu / \sigma$)& 0.04 / 0.00&  0.07 / 0.00 &  \textbf{0.29 / 0.00} &0.09 / 0.00  & 0.12 / 0.00\\
                                &SD ($\mu / \sigma$)& 1.89 / 3.97 & \textbf{1.58 / 4.11} &  \textbf{1.22 / 3.32} & \textbf{0.98 / 0.65} & 1.43 / 8.02\\
\hline
\multirow{3}{*}{Fast-LIO2~\cite{xu2022fast}} 
                              & ATE ($\mu / \sigma$) & 1.15 / 0.67 & \textbf{1.51 / 0.82} & 9.77 / 13.51 & \textbf{0.87 / 0.30} & 2.50 / 0.40\\
                              & RPE ($\mu / \sigma$)  & \textbf{0.01 / 0.00}  &  \textbf{0.01 / 0.00}  & 1.13 / 0.40 & \textbf{0.00 / 0.00} & \textbf{0.04 / 0.00}\\
                              &SD ($\mu / \sigma$)& 1.13 / 0.70 & 1.67 / 4.09 &  1.95 / 6.40 & 1.03 / 0.67 & \textbf{1.02 / 0.58}\\
\hline
\multirow{3}{*}{DLIO~\cite{chen2022dlio}} 
                              & ATE ($\mu / \sigma$) & \textbf{1.14 / 0.49} & 2.64 / 1.14 & \textbf{6.93 / 8.81} & 3.29 / 2.47 & \textbf{1.76 / 1.65} \\
                              & RPE ($\mu / \sigma$)& 0.04 / 0.00 & 0.03 / 0.00  & \textbf{0.29 / 0.00} & 0.04 / 0.00 & 0.05 / 0.00 \\
                              &SD ($\mu / \sigma$)& \textbf{1.10 / 0.50}  &  2.19 / 6.34 & 1.45 / 3.94 & 1.19 / 3.19 & 1.12 / 2.05 \\
\hline
\multirow{3}{*}{RTAB~\cite{rtab}} 
                               & ATE ($\mu / \sigma$) & 12.36 / 6.85 &46.06 / 22.23 & \multirow{3}{*}{0} & 3.25 / 2.15 & 5.63 / 2.64 \\
                               & RPE ($\mu / \sigma$)  & 0.03 / 0.00  & 0.08 / 0.00 && 0.09 / 0.00 & 0.09 / 0.00\\
                               &SD ($\mu / \sigma$)& 0.24 / 4.37 & 2.05 / 4.58 && 0.94 / 0.89 & 0.46 / 8.37 \\
\hline
\multirow{3}{*}{ORB-SLAM3~\cite{orbslam}} 
                               & ATE ($\mu / \sigma$) & 23.37 / 9.99 &21.93 / 10.31& \multirow{3}{*}{0} & 54.74 / 35.04 & 53.98 / 24.86\\
                               & RPE ($\mu / \sigma$)  & 0.04 / 0.00  & 0.05 / 0.00 & & 0.05 / 0.00 & 0.08 / 0.00\\
                               &SD ($\mu / \sigma$)& 0.88 / 2.72 & 1.89 / 6.99 && 0.13 / 1.58 & 0.03 / 0.15\\
\hline
\vspace{-5.5pt}
\end{tabular}
\textbf{$\mu \rightarrow$} mean, $\sigma \rightarrow $ standard deviation. \textbf{V.Illum.} $\rightarrow$ Varying illumination. \textbf{HD Obs. \& F-Spar} $\rightarrow$ High dimensional observations and feature sparsity.
\label{tab: indoorstat}
\vspace*{-1.25\baselineskip}
\end{table*}

\subsection{Feature Sparsity and Density}\label{featDensities}
This section relates to RQ2, where we seek to examine whether the distributions of feature points (e.g., clustered, sparse, or evenly distributed) might affect the navigation precision and the quality of the generated map. To achieve that, we evaluate the spatial distribution of feature points and correlate it with the per-point ATE and RPE.

We employed a voxel grid approach to compute the feature density of the point clouds representing the scene. We define a voxel grid to compute the feature distribution of the given point cloud data from each scene.
Formally, let $\mathbf{O} = [\mathbf{o}_1, \mathbf{o}_2, \ldots, \mathbf{o}_m]^\top$ represent the occupied cells in the scene consisting of $m$ points, $\mathbf{o}_j \in \mathbb{R}^3$.
To compute the feature density, we divide the bounding box of the point cloud into a regular grid of voxels with specified voxel size $\mathbf{v}$. For each point \(\mathbf{o}_i = (\mathbf{x_{i}}, \mathbf{y_{i}}, \mathbf{z_{i}})\), we determine its corresponding voxel by computing, \(\mathbf{v}_i = \mathbf{\left( (x_i - x_{\min})/{v}, (y_i - y_{\min})/{v}, (z_i - z_{\min})/{v} \right)}\),
where \( \mathbf{(x_{\min}, y_{\min}, z_{\min})} \) are the minimum coordinates of the occupied cell arrays. The feature density \( \rho(\mathbf{v}) \) for each voxel \( \mathbf{v} \) is then computed as 
 $\rho(\mathbf{v}) = \sum_{i=1}^n \mathbf{1}_{\{\mathbf{v}_i = \mathbf{v}\}}$, where \( \mathbf{1}_{\{\mathbf{v}_i = \mathbf{v}\}} \) is an indicator function that evaluates whether the point \(\mathbf{o}_i\) falls within the voxel \(\mathbf{v}\). \( \mathbf{v}_i \) is the voxel index of point \( \mathbf{o}_i \), and \( n \) is the total number of points in the occupied cell arrays.

To establish the correlation between the feature density and the errors (ATE \& RPE), first, we map the trajectory points to feature density bins. For each point, e.g., $\hat{\mathbf{t}}_i = (\hat{x}_i, \hat{y}_i, \hat{z}_i)$, we find the bin index \((x_{\text{bin}, i}, y_{\text{bin}, i}, z_{\text{bin}, i})\) for each coordinate with the corresponding edges \((\hat{x}_i \text{ in } \mathbf{x}_\text{edges}, \hat{y}_i \text{ in } \mathbf{y}_\text{edges}, \hat{z}_i \text{ in } \mathbf{z}_\text{edges})\). Thereafter, we assign the feature density at the point as:
$ \mathbf{d}_i =  \rho(\textbf{x}_{\text{bin}, i}, \textbf{y}_{\text{bin}, i}, \textbf{z}_{\text{bin}, i})~\text{if bin indices are valid}$, and 0 otherwise.
\( \mathbf{d} = [d_1, d_2, \ldots, d_n]\) is the feature density at the trajectory points, \(\rho(.)\) is the feature density of the voxel that contains the point. For simplicity, we assume that out-of-bounds points have zero features.
Finally, we compute the 2D Pearson correlation between the errors and the feature densities \(\mathbf{d}\) as \(\rho_{\text{ATE}, \mathbf{d}} = \text{corr}(\text{ppATE}, \mathbf{d})\) and \(\rho_{\text{RPE}, \mathbf{d}} = \text{corr}(\text{ppRPE}, \mathbf{d})\).
\begin{figure*}[htp]
 \centering
 \subfigure[Leo-Str]{\includegraphics[scale=0.21]{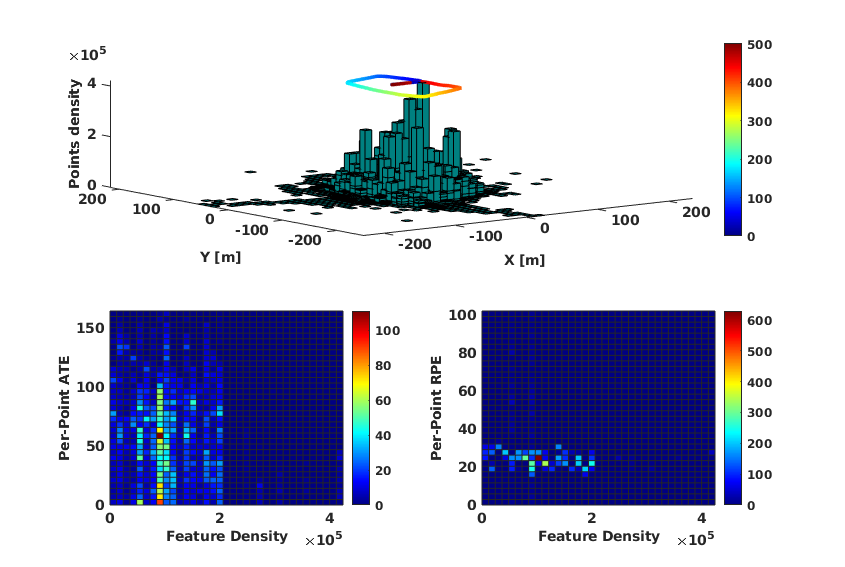}}
 \subfigure[MU-Hall]{\includegraphics[scale=0.25]{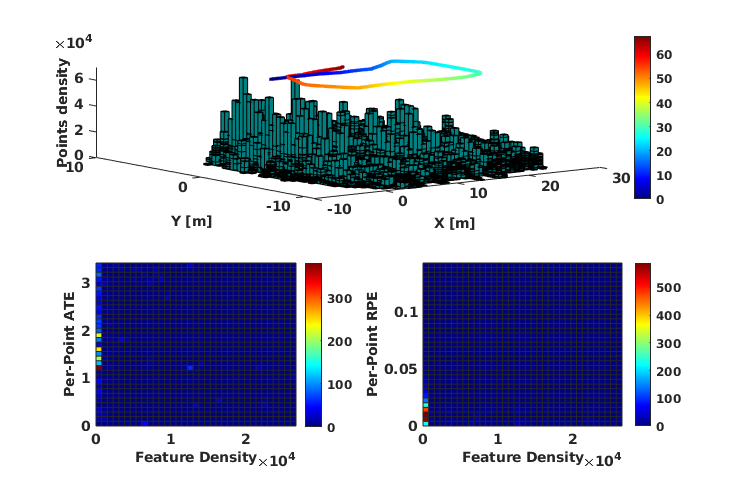}}
  \subfigure[Leo-For]{\includegraphics[scale=0.22]{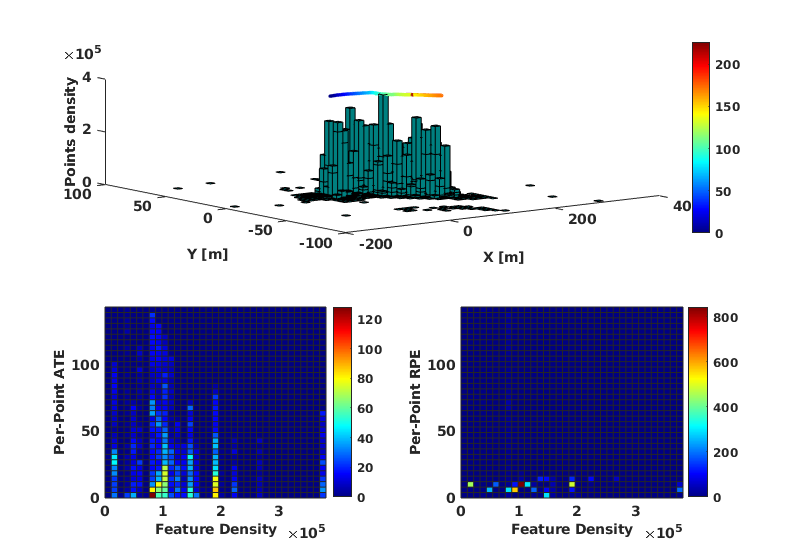}}
 \vspace{-5pt}
 \caption{Correlation between feature point distributions (clustered, sparse, and evenly distributed) and per-point trajectory errors (ppATE and ppRPE). The top section shows feature density in the reconstructed map, with robot trajectories coloured by distance (blue-start and red-end). The bottom section highlights the correlation between feature density and trajectory errors, with colour intensity representing error frequency.}
 \vspace*{-0.8\baselineskip}
\label{fig:ppATE-RPE}
\end{figure*}

Our results in Fig.\ref{fig:ppATE-RPE} showed a complex non-linear relationship between the errors and the feature densities, contrary to the common assumption that more features improve perception and SLAM accuracy.  
In Fig.\ref{fig:ppATE-RPE}(c) for example, the highest ATE errors occurred within a feature density range of $1.0 \times 10^{5}$ to $2.1 \times 10^{5}$, with significant errors around $1$ to $5 m$. This indicates that both sparse and dense feature environments can challenge SLAM algorithms.
RPE errors also correlated with mid-range feature densities, with significant values from $0.5 \times 10^{5}$ to $1.5 \times 10^{5}$ feature density.

\subsection{Supervised Learning Algorithms Benchmark}\label{sec:detection}
We address RQ3 using the EnvoDat by evaluating three pre-trained object detector models: YOLOv8 \cite{yolov8_ultralytics}, Fast R-CNN \cite{7410526}, and Detectron2 \cite{wu2019detectron2}. We trained these models on the annotated RGB images drawn across all the scenes in the EnvoDat, which exhibit the characteristics outlined in RQ3. 
For consistency, we trained all the models with equal hyperparameter settings (e.g., learning rate $= 0.00025$, batch size $= 16$, epochs $=150$), with a 70\% train, 20\% validation, and 10\% test split ratios. We evaluated their performances based on accuracy, precision, and efficiency.

For accuracy, we used the mean average precision (mAP) metric at multiple intersections over union (IoU) thresholds, $0.5$ and $0.5 - 0.95$. Formally, $\text{mAP}_{0.5} = \frac{1}{N} \sum_{i=1}^{N} \text{AP}_{i}^{(\text{IoU}=0.5)}$ and $\text{mAP}_{(0.5:0.95)} = \frac{1}{N} \sum_{i=1}^{N} \frac{1}{10} \sum_{t=0.5}^{0.95} \text{AP}_{i}^{(\text{IoU}=t)}$, where \(N\) is the total number of object classes, and \(\text{AP}_{i}\) is the average precision for class \(i\). 
\vspace{-2.5pt}
\begin{table}[htbp]
    \centering
    \caption{Performance Comparison of Object Detection Models}
    \vspace{-7.5pt}
    \begin{tabular}{lcccccc}
        \toprule
        \textbf{Metric} & \textbf{YOLOv8} & \textbf{Detectron2} & \textbf{Fast R-CNN} \\ 
        \hline
        \(mAP_{0.5} \)(\%)           &\textbf{71.85}       &50.10      &60.40 \\ 
        \(mAP_{(0.5:0.95)}\)(\%)     & \textbf{61.00}      &41.70      &42.10 \\ 
        Precision(\%)             & \textbf{72.85}       &50.01      &50.40\\ 
        Recall(\%)                 & \textbf{68.90}       &48.65      &48.50\\ 
        F1-score                  & \textbf{0.709}       &0.54       &0.54 \\ 
         Inference Time (ms)       &\textbf{8.75}        &18.20      &20.20\\ 
        Frames Per Second          &\textbf{114.29}      &54.95      &49.50 \\
        Memory Usage (GB)           &\textbf{4.71}        &16.10      &16.70 \\ 
        \bottomrule
    \end{tabular}
    \label{tab:model_performance}
    \vspace*{-1.25\baselineskip}
\end{table}
For precision, we utilised the precision and recall metrics, similar to \cite{mortimer2024goose} and \cite{doi:10.1177/02783649241242136}. 
To balance these two metrics, we also computed the F1-score, the harmonic mean of precision and recall. 
For efficiency, we evaluated the models in terms of inference time (\(t_\text{inf}\)), speed (frame per second i.e., \(\text{FPS} = 1/t_{\text{inf}}\)), and GPU memory usage.

Table~\ref{tab:model_performance} shows the models' performance on EnvoDat. The environment heterogeneity had a noticeable impact on their results. Nevertheless, YOLOv8 outperformed other models across most metrics, particularly in mAP and inference speed. Detectron2 and Fast R-CNN lagged, especially in inference time and memory usage. Thus, YOLOv8 can be considered a more efficient choice for real-time applications.

\section{Conclusion}
We introduced EnvoDat, a large-scale multimodal dataset for advancing robotic autonomy, spatial awareness and semantic reasoning in heterogeneous environments. EnvoDat covers a range of indoor, outdoor, and subterranean environments with distinct geometric and feature characteristics. We benchmarked several SOTA SLAM and supervised learning algorithms on EnvoDat, showing how real-world conditions influence their performance in challenging environments, including multi-degraded, dynamic, and feature-sparse areas. We plan to expand our dataset by capturing more sequences, adding new scenes, and providing more fine-grained layered polygon-based annotations to accelerate the development of real-time ecocentric perception and SLAM algorithms.

\addtolength{\textheight}{-1cm}   

    \bibliographystyle{ieeetr}
    \bibliography{references}

\begin{thebibliography}{10}

\bibitem{Stachniss2016}
C.~Stachniss, J.~J. Leonard, and S.~Thrun, {\em Simultaneous Localization and Mapping}, pp.~1153--1176.
\newblock Cham: Springer International Publishing, 2016.

\bibitem{slam1}
C.~Cadena, L.~Carlone, H.~Carrillo, Y.~Latif, D.~Scaramuzza, J.~Neira, I.~Reid, and J.~J. Leonard, ``Past, present, and future of simultaneous localization and mapping: Toward the robust-perception age,'' {\em IEEE Transactions on Robotics}, vol.~32, no.~6, pp.~1309--1332, 2016.

\bibitem{Nwankwo_2023}
L.~Nwankwo and E.~Rueckert, {\em Understanding Why SLAM Algorithms Fail in Modern Indoor Environments}, p.~186–194.
\newblock Springer Nature Switzerland, 2023.

\bibitem{kaveti2023challenges}
P.~Kaveti, A.~Gupta, D.~Giaya, M.~Karp, C.~Keil, J.~Nir, Z.~Zhang, and H.~Singh, ``Challenges of indoor slam: A multi-modal multi-floor dataset for slam evaluation,'' in {\em 2023 IEEE 19th International Conference on Automation Science and Engineering (CASE)}, pp.~1--8, IEEE, 2023.

\bibitem{Zhao2024CVPR}
S.~Zhao, Y.~Gao, T.~Wu, D.~Singh, R.~Jiang, H.~Sun, M.~Sarawata, Y.~Qiu, W.~Whittaker, I.~Higgins, Y.~Du, S.~Su, C.~Xu, J.~Keller, J.~Karhade, L.~Nogueira, S.~Saha, J.~Zhang, W.~Wang, C.~Wang, and S.~Scherer, ``Subt-mrs dataset: Pushing slam towards all-weather environments,'' in {\em Proceedings of the IEEE/CVF Conference on Computer Vision and Pattern Recognition (CVPR)}, pp.~22647--22657, June 2024.

\bibitem{10415477}
Y.~Liu, Y.~Fu, M.~Qin, Y.~Xu, B.~Xu, F.~Chen, B.~Goossens, P.~Z. Sun, H.~Yu, C.~Liu, L.~Chen, W.~Tao, and H.~Zhao, ``Botanicgarden: A high-quality dataset for robot navigation in unstructured natural environments,'' {\em IEEE Robotics and Automation Letters}, vol.~9, no.~3, pp.~2798--2805, 2024.

\bibitem{kitti}
A.~Geiger, P.~Lenz, and R.~Urtasun, ``Are we ready for autonomous driving? the kitti vision benchmark suite,'' in {\em 2012 IEEE Conference on Computer Vision and Pattern Recognition}, pp.~3354--3361, 2012.

\bibitem{roboCar}
W.~Maddern, G.~Pascoe, C.~Linegar, and P.~Newman, ``{1 Year, 1000km: The Oxford RobotCar Dataset},'' {\em The International Journal of Robotics Research (IJRR)}, vol.~36, no.~1, pp.~3--15, 2016.

\bibitem{tumVi}
S.~Klenk, J.~Chui, N.~Demmel, and D.~Cremers, ``Tum-vie: The tum stereo visual-inertial event dataset,'' in {\em 2021 IEEE/RSJ International Conference on Intelligent Robots and Systems (IROS)}, pp.~8601--8608, 2021.

\bibitem{9968057}
L.~Zhang, M.~Helmberger, L.~F.~T. Fu, D.~Wisth, M.~Camurri, D.~Scaramuzza, and M.~Fallon, ``Hilti-oxford dataset: A millimeter-accurate benchmark for simultaneous localization and mapping,'' {\em IEEE Robotics and Automation Letters}, vol.~8, no.~1, pp.~408--415, 2023.

\bibitem{hilti}
M.~Helmberger, K.~Morin, B.~Berner, N.~Kumar, G.~Cioffi, and D.~Scaramuzza, ``The hilti slam challenge dataset,'' {\em IEEE Robotics and Automation Letters}, vol.~7, pp.~1--8, 07 2022.

\bibitem{tum}
J.~Sturm, N.~Engelhard, F.~Endres, W.~Burgard, and D.~Cremers, ``A benchmark for the evaluation of rgb-d slam systems,'' in {\em Proc. of the International Conference on Intelligent Robot Systems (IROS)}, Oct. 2012.

\bibitem{wild-places}
J.~Knights, K.~Vidanapathirana, M.~Ramezani, S.~Sridharan, C.~Fookes, and P.~Moghadam, ``Wild-places: A large-scale dataset for lidar place recognition in unstructured natural environments,'' in {\em 2023 IEEE International Conference on Robotics and Automation (ICRA)}, pp.~11322--11328, 2023.

\bibitem{sun}
S.~Song, S.~P. Lichtenberg, and J.~Xiao, ``Sun rgb-d: A rgb-d scene understanding benchmark suite,'' in {\em 2015 IEEE Conference on Computer Vision and Pattern Recognition (CVPR)}, pp.~567--576, 2015.

\bibitem{7780719}
M.~Cordts, M.~Omran, S.~Ramos, T.~Rehfeld, M.~Enzweiler, R.~Benenson, U.~Franke, S.~Roth, and B.~Schiele, ``The cityscapes dataset for semantic urban scene understanding,'' in {\em 2016 IEEE Conference on Computer Vision and Pattern Recognition (CVPR)}, pp.~3213--3223, 2016.

\bibitem{locdata}
D.~Lee, S.~Ryu, S.~Yeon, Y.~Lee, D.~Kim, C.~Han, Y.~Cabon, P.~Weinzaepfel, N.~Guérin, G.~Csurka, and M.~Humenberger, ``Large-scale localization datasets in crowded indoor spaces,'' in {\em 2021 IEEE/CVF Conference on Computer Vision and Pattern Recognition (CVPR)}, pp.~3226--3235, 2021.

\bibitem{complexUrban}
J.~Jeong, Y.~Cho, Y.~sik Shin, H.~Roh, and A.~Kim, ``Complex urban dataset with multi-level sensors from highly diverse urban environments,'' {\em The International Journal of Robotics Research}, vol.~38, pp.~642 -- 657, 2019.

\bibitem{cadc}
M.~Pitropov, D.~E. Garcia, J.~Rebello, M.~Smart, C.~Wang, K.~Czarnecki, and S.~Waslander, ``Canadian adverse driving conditions dataset,'' {\em The International Journal of Robotics Research}, vol.~40, no.~4-5, pp.~681--690, 2021.

\bibitem{10542164}
C.~Yao, Y.~Ge, G.~Shi, Z.~Wang, N.~Yang, Z.~Zhu, H.~Wei, Y.~Zhao, J.~Wu, and Z.~Jia, ``Tail: A terrain-aware multi-modal slam dataset for robot locomotion in deformable granular environments,'' {\em IEEE Robotics and Automation Letters}, vol.~9, no.~7, pp.~6696--6703, 2024.

\bibitem{9197298}
G.~Kim, Y.~S. Park, Y.~Cho, J.~Jeong, and A.~Kim, ``Mulran: Multimodal range dataset for urban place recognition,'' in {\em 2020 IEEE International Conference on Robotics and Automation (ICRA)}, pp.~6246--6253, 2020.

\bibitem{10530418}
A.~Zhang, C.~Eranki, C.~Zhang, J.-H. Park, R.~Hong, P.~Kalyani, L.~Kalyanaraman, A.~Gamare, A.~Bagad, M.~Esteva, and J.~Biswas, ``Toward robust robot 3-d perception in urban environments: The ut campus object dataset,'' {\em IEEE Transactions on Robotics}, vol.~40, pp.~3322--3340, 2024.

\bibitem{doi:10.1177/02783649241242136}
M.~Jung, W.~Yang, D.~Lee, H.~Gil, G.~Kim, and A.~Kim, ``Helipr: Heterogeneous lidar dataset for inter-lidar place recognition under spatiotemporal variations,'' {\em The International Journal of Robotics Research}, vol.~0, no.~0, p.~02783649241242136, 0.

\bibitem{Warburg_2020_CVPR}
F.~Warburg, S.~Hauberg, M.~Lopez-Antequera, P.~Gargallo, Y.~Kuang, and J.~Civera, ``Mapillary street-level sequences: A dataset for lifelong place recognition,'' in {\em Proceedings of the IEEE/CVF Conference on Computer Vision and Pattern Recognition (CVPR)}, June 2020.

\bibitem{Characterizing19}
S.~Saeedi, E.~D.~C. Carvalho, W.~Li, D.~Tzoumanikas, S.~Leutenegger, P.~H.~J. Kelly, and A.~J. Davison, ``Characterizing visual localization and mapping datasets,'' in {\em International Conference on Robotics and Automation (ICRA)}, pp.~6699--6705, IEEE, 2019.

\bibitem{9809788}
L.~Gao, Y.~Liang, J.~Yang, S.~Wu, C.~Wang, J.~Chen, and L.~Kneip, ``Vector: A versatile event-centric benchmark for multi-sensor slam,'' {\em IEEE Robotics and Automation Letters}, vol.~7, no.~3, pp.~8217--8224, 2022.

\bibitem{McCormac_2017_ICCV}
J.~McCormac, A.~Handa, S.~Leutenegger, and A.~J. Davison, ``Scenenet rgb-d: Can 5m synthetic images beat generic imagenet pre-training on indoor segmentation?,'' in {\em Proceedings of the IEEE International Conference on Computer Vision (ICCV)}, Oct 2017.

\bibitem{Yin2023GroundChallengeAM}
J.~Yin, H.~Yin, C.~Liang, and Z.~Zhang, ``Ground-challenge: A multi-sensor slam dataset focusing on corner cases for ground robots,'' {\em 2023 IEEE International Conference on Robotics and Biomimetics (ROBIO)}, pp.~1--5, 2023.

\bibitem{citrusFarm}
H.~Teng, Y.~Wang, X.~Song, and K.~Karydis, ``Multimodal dataset for localization, mapping and crop monitoring in citrus tree farms,'' in {\em International Symposium on Visual Computing}, pp.~571--582, 2023.

\bibitem{envodat}
L.~Nwankwo, B.~Ellensohn, V.~Dave, P.~Hofer, J.~Forstner, M.~Villneuve, R.~Galler, and E.~Rueckert, ``Envodat: A large-scale multisensory dataset for robotic spatial awareness and semantic reasoning in heterogeneous environments.''
\newblock Project Website: \url{https://linusnep.github.io/EnvoDat/}.

\bibitem{articleGPS}
A.~Jacobson, F.~Zeng, D.~Smith, N.~Boswell, T.~Peynot, and M.~Milford, ``What localizes beneath: A metric multisensor localization and mapping system for autonomous underground mining vehicles,'' {\em Journal of Field Robotics}, vol.~38, 08 2020.

\bibitem{ros}
M.~Quigley, K.~Conley, B.~Gerkey, J.~Faust, T.~Foote, J.~Leibs, R.~Wheeler, and A.~Ng, ``Ros: an open-source robot operating system,'' vol.~3, 01 2009.

\bibitem{nwankwo2023romr}
L.~Nwankwo, C.~Fritze, K.~Bartsch, and E.~Rueckert, ``Romr: A ros-based open-source mobile robot,'' {\em HardwareX}, vol.~14, p.~e00426, 2023.

\bibitem{glim}
K.~Koide, M.~Yokozuka, S.~Oishi, and A.~Banno, ``Glim: 3d range-inertial localization and mapping with gpu-accelerated scan matching factors,'' {\em Robotics and Autonomous Systems}, vol.~179, p.~104750, Sept. 2024.

\bibitem{roboflow}
B.~Dwyer, J.~Nelson, T.~Hansen, and et. al., ``Roboflow (version 1.0) [software],'' {\em Available from \url{https://roboflow.com}. computer vision.}, 2024.

\bibitem{rtab}
M.~Labbé and F.~Michaud, ``Rtab-map as an open-source lidar and visual simultaneous localization and mapping library for large-scale and long-term online operation,'' {\em Journal of Field Robotics}, vol.~36, no.~2, pp.~416--446, 2019.

\bibitem{orbslam}
C.~Campos, R.~Elvira, J.~J.~G. Rodríguez, J.~M. M.~Montiel, and J.~D.~Tardós, ``Orb-slam3: An accurate open-source library for visual, visual–inertial, and multimap slam,'' {\em IEEE Transactions on Robotics}, vol.~37, no.~6, pp.~1874--1890, 2021.

\bibitem{hdl}
K.~Koide, M.~Jun, and E.~Menegatti, ``A portable 3d lidar-based system for long-term and wide-area people behavior measurement,'' 2019.

\bibitem{xu2022fast}
W.~Xu, Y.~Cai, D.~He, J.~Lin, and F.~Zhang, ``Fast-lio2: Fast direct lidar-inertial odometry,'' {\em IEEE Transactions on Robotics}, vol.~38, no.~4, pp.~2053--2073, 2022.

\bibitem{chen2022dlio}
K.~Chen, R.~Nemiroff, and B.~T. Lopez, ``Direct lidar-inertial odometry: Lightweight lio with continuous-time motion correction,'' {\em 2023 IEEE International Conference on Robotics and Automation (ICRA)}, pp.~3983--3989, 2023.

\bibitem{yolov8_ultralytics}
G.~Jocher, A.~Chaurasia, and J.~Qiu, ``Ultralytics yolov8,'' 2023.

\bibitem{7410526}
R.~Girshick, ``Fast r-cnn,'' in {\em 2015 IEEE International Conference on Computer Vision (ICCV)}, pp.~1440--1448, 2015.

\bibitem{wu2019detectron2}
Y.~Wu, A.~Kirillov, F.~Massa, W.-Y. Lo, and R.~Girshick, ``Detectron2.'' \url{https://github.com/facebookresearch/detectron2}, 2019.

\bibitem{mortimer2024goose}
P.~Mortimer, R.~Hagmanns, M.~Granero, T.~Luettel, J.~Petereit, and H.-J. Wuensche, ``{The GOOSE Dataset for Perception in Unstructured Environments},'' in {\em {Proceedings of IEEE International Conference on Robotics and Automation (ICRA)}}, 2024.

\end{thebibliography}

\end{document}